\pdfoutput=1

\documentclass{article}
\usepackage[preprint]{neurips_2024}

\usepackage[utf8]{inputenc} 
\usepackage[T1]{fontenc}    
\usepackage{hyperref}       
\usepackage{url}            
\usepackage{booktabs}       
\usepackage{amsfonts}       
\usepackage{nicefrac}       
\usepackage{microtype}      
\usepackage{xcolor}         
\usepackage{times}
\usepackage{latexsym}
\usepackage{inconsolata}
\usepackage{soul}
\usepackage{graphicx}
\usepackage{amsmath, amssymb}
\usepackage{amsthm}
\usepackage{algorithm}
\usepackage{algorithmic}
\usepackage{multirow}
\usepackage{color}
\usepackage{bm}
\usepackage{bbm}
\usepackage{natbib}
\usepackage{enumitem}

\title{Multidimensional Rubric-oriented Reward Model Learning via Geometric Projection Reference Constraints}

\author{Yongnan Jin\thanks{\quad These authors contributed equally to this work.}, Xurui Li$^*$\thanks{\quad Corresponding authors.}, \hspace{0.1em} Feng Cao, Liucun Gao$^{\dag}$, Juanjuan Yao$^{\dag}$\\
        Shanghai Mingpin Medical Data Technology Co., Ltd., China \\ }

\begin{document}

\maketitle

\begin{abstract}
The integration of large language models (LLMs) into medical practice offers transformative potential, yet their real-world clinical applicability remains constrained by critical alignment issues: (1) a misalignment between static evaluation benchmarks and the dynamic cognitive demands of clinical practice, (2) challenges in adapting to continuously evolving, multi-source medical standards, and (3) the limited capacity of conventional reward models to reflect nuanced, multi-dimensional medical quality criteria. To overcome these limitations, we introduce MR-RML (Multidimensional Rubric-oriented Reward Model Learning) with GPRC (Geometric Projection Reference Constraints)-a novel alignment framework that structured medical standards into a multi-perspective matrix to guide both data generation and model optimization. Our approach introduces three key innovations: (1) a medical standard system that embeds domain-specific guidelines throughout the training pipeline; (2) an independent multi-dimensional reward model that decomposes evaluation criteria, transitioning from rule-based or LLM-based scoring to internalized reward modeling for better evaluation performance; and (3) geometric projection reference constraints that translate clinical cognitive logic into mathematical regularization, aligning scoring gradients with clinical reasoning and facilitating training with synthetically generated data. Extensive evaluations on the authoritative medical benchmark Healthbench demonstrate that our method significantly boosts the performance of the base Qwen-32B model, with improvements of 45\% on the full subset and 85\% on the hard subset. It achieves state-of-the-art results among open-source LLMs, scoring 62.7 (full) and 44.7 (hard), while also surpassing the majority of closed-source models.

\end{abstract}

\section{Introduction}
The application of large language models in the medical field has gradually expanded from basic disease QA to complex scenarios such as auxiliary diagnosis, medication recommendations, and chronic disease management \citep{liu2025application, zhang2024chatbot, lievin2024can}. However, the error rate of models in real medical environments remains unacceptably high. The fundamental cause lies in three inherent flaws of the traditional training paradigm in medical scenarios, which severely restrict the medical utility and trustworthiness of medical LLMs:

\begin{itemize}[left=0pt]
\item \textbf{Misalignment Between Evaluation and Medical Cognition.}
Static evaluation systems fail to match the dynamic cognitive needs of medical scenarios. Medical cognitive priorities vary significantly, for example, emergency scenarios require prioritizing emergency level judgment and timeliness of risk prompting, while chronic disease management focuses on completeness of follow-up recommendations and guidance for patient adherence. Existing evaluation benchmarks mostly adopt fixed weight assignment models, failing to reflect these differences. Meanwhile, evaluation dimensions are overly simplified: mainstream benchmarks such as MedQA \citep{roy2024beyond} use multiple-choice questions to assess medical fact memorization, but cannot cover medical-critical requirements like compliance, communication quality, and reasoning logic. Although HealthBench \citep{arora2025healthbench} advances scenario-based evaluation, it only provides datasets without evaluation-training closed-loop, limiting its ability to guide model optimization or generalize to specific industry standards.

\item \textbf{Bottlenecks in Medical Standard Adaption for LLMs.}
Medical standards are characterized by dynamic updates, and multi-source heterogeneity-different regions, scenarios, and departments have exclusive medical guidelines, with ethical norms revised annually. Existing solutions struggle to balance adaptability and cost efficiency. Extending Reinforcement Learning with Verifiable Rewards (RLVR) \citep{su2025crossing} to real-world tasks is challenging, as evaluation relies on nuanced multi-criteria judgments rather than binary correctness. Recent works have extend RLVR beyond strictly verifiable domains by using structured rubrics as reward signals for multi-dimensional/subjective tasks. \citep{huang2025reinforcement} constructs a large-scale rubric library but lacks deep integration with authoritative medical standards, and its rule-based scoring has low reusability. RaR \citep{gunjal2025rubrics} framework implements multi-dimensional evaluation via LLM-generated instance-specific rubrics, but suffers from high real-time scoring costs, task-dependent standards, and over-reliance on LLM-as-Judge \citep{jang2025instajudge, son2024llm} during Reinforcement Learning (RL). A common shortcoming is the absence of an independently trained reward model, leading to inconsistent evaluation, poor generalization, or excessive training costs.

\item \textbf{Dilemmas in Multi-Dimensional Reward Model Training.}
Traditional Reward Model Learning (RML) \citep{nika2024reward} face three insurmountable limitations in medicine: extremely high annotation costs requiring senior experts, single-scalar output lacking dimension decomposition (failing regulatory and trust requirements), and absence of mathematical constraints (resulting in irrational scoring gradients and weak generalization). Existing loss functions optimize only a few dimensions, unable to adapt to the multi-orthogonal requirements of medical accuracy, safety, and ethical compliance, while ignoring semantic space geometric relationships between samples.
\end{itemize}
 
To address these flaws, we propose the MR-RML (Multidimensional Rubric-oriented Reward Model Learning) via GPRC (Geometric Projection Reference Constraints), which reconstructs the alignment pathway of medical LLMs with three core innovations:
\begin{itemize}[left=0pt]
     \item End-to-end standard alignment innovation: A scenario-dimension collaborative calibration matrix integrates medical cognitive axis standards into data generation and model training for SFT, RML and RL, resolving the disconnection between evaluation and standard-aligned training.
    \item Reinforcement learning scoring paradigm innovation: Medical standard knowledge is internalized into a multi-dimensional reward model, enabling refined decomposition and controllable optimization of evaluation dimensions. This shifts the RLVR or real-time rubric-level scoring for RL to "multi-dimensional internalized reward modeling", improving professionalism, consistency, and generalization at lower labor-costs.
    \item Reward model innovation: From "single scalar aggregated reward" to "geometric projection reference-constrained reward", medical cognitive logic is transformed into mathematical constraints. Sample references and geometric projection consistency ensure scoring gradients align with medical thinking, reducing reliance on high-quality samples and enabling training with large-scale synthetic data.
\end{itemize}
This study aims to provide a "medical cognition-oriented, standard-extensible, and cost-controllable" development solution for medical LLMs, promoting medical AI from technically feasible to medically trustworthy. Extensive evaluations on the authoritative medical benchmark Healthbench demonstrate that our method delivers significant improvements against base LLM model and achieve the SOTA performance, as well as surpassing the majority of closed-source models.\footnote{\footnotesize{\url{https://www.modelscope.cn/models/dazhuanjia/Shanzhi-M1}}}

\section{Related Work}
\subsection{Paradigm Evolution of Medical LLM Evaluation and Alignment}
Medical LLM evaluation has evolved from "knowledge assessment" to "capability evaluation" and "scenario adaptation," but core limitations persist. Early methods (e.g. MedQA) \citep{roy2024beyond} focus on medical factual knowledge via multiple-choice questions. They are easy to quantify but fail to evaluate medical-critical capabilities like reasoning, communication, and compliance, leading to a disconnect from real needs. Subsequent benchmarks (HealthBench) \citep{arora2025healthbench} introduce scenario-based tasks with refined evaluation rules. However, they lack a "evaluation-training" closed-loop, limiting generalization across scenarios and dimensions. Reinforcement Learning with Verifiable Rewards (RLVR) shows potential in medical alignment, while extending it to real-world tasks is challenging, as evaluation relies on nuanced multi-criteria judgments rather than binary correctness. Recent structured rubric systems address multi-dimensional evaluation by rule-based or LLM-based rubric-level on-training scoring, which is either static or high-cost: (1) Rubicon-preview \citep{huang2025reinforcement}: A rubric library with tens of thousands of standards, extending verifiable reward RL to open domains. Limitations include general-task-oriented dimensions, poor rule reusability, and reliance on pre-defined rules (inconsistent evaluation). (2) RaR \citep{gunjal2025rubrics} implements evaluation using instance-specific rubrics generated by LLMs, transforming rubrics from mere evaluation tools into reward mechanisms and bridging the gap between verifiable rewards and preference ranking. This LLM-as-Judge approach is also adopted in \citep{ye2025self, xie2025auto, jacob2025qa}, with additional rubrics learning processes, etc. However, they have several core limitations: first, rubric generation depends on real-time LLM-based scoring, which is costly and lacks sufficient compliance depth for medical scenarios; second, rubric standards are tightly coupled with specific tasks, necessitating re-generation when new evaluation dimensions are added; third, it over-relies on LLM-as-Judge during reinforcement learning  training, leading to extremely high training costs and imposing strict requirements on both the consistency of the scoring LLM and the quality of training data standards.

Our method MR-RML offers three advantages: medical cognitive axes aligned with authoritative medical standards, an independent multi-dimensional reward model (reducing costs and improving consistency), and geometric reference constraints enabling a synthetic data-driven "evaluation-training" closed-loop.

\subsection{Adaptation Challenges of Reward Models in Vertical Domains}
As core alignment components, reward models are mature in general domains but face three challenges in medicine: professionalism, interpretability, and annotation cost.\citep{zhong2025comprehensive, li2025generalist, ji2025survey, wu2025sailing, zang2025internlm}
\begin{itemize} [left=0pt]
    \item Professionalism: Medical evaluation involves orthogonal dimensions (accuracy, safety, compliance, empathy) with dynamic scenario-specific weights, but traditional RMs output single scalars.
    \item Interpretability: Medical applications require decision transparency, but traditional "black-box" RMs provide no detailed explanation for overall scores.
    \item Cost: High-quality medical preference data requires expert annotation; even RLAIF faces data quality and compliance issues.
\end{itemize}

Mainstream RMs are mismatched with medical needs. Skywork-Reward series \citep{liu2024skywork, liu2025skywork} uses Bradley-Terry ranking loss and large-scale preference pairs, but single-dimensional ranking cannot capture medical multi-dimensionality, and high-quality samples are hard to obtain. POLAR \citep{doupre} reconstructs RMs as "policy difference measurers," but suffers from single-dimensional evaluation, weak correlation between policy distance and medical quality, and high reference trajectory costs.

To address these limitations, MR-RML proposes multi-medical cognitive axis geometric constraint loss, achieving adaptation via "dimension decomposition + logical constraints". Compared to existing solutions, MR-RML’s medical cognitive axes align with medical standards, the independent reward model balances consistency and cost, and full-link closed-loop solves generalization issues.

\section{Methodology}
\subsection{Construction of the 3D Medical Standard System: Dimensions-Scenarios-Disciplines}
A "Dimensions-Scenarios-Disciplines" 3D medical standard system is constructed to establish a systematic, quantifiable medical quality benchmark, guiding subsequent data generation and model training.

\subsubsection{Definition of Core Variables}
\begin{itemize} [left=0pt]
    \item $L$ is the number of core dimensions for medical evaluation. Aligned with domain experts and authoritative standards (e.g., \textit{Information Content Quality}, \textit{User Intent Recognition and Triage} and \textit{Clinical Assessment and Diagnostic Reasoning}), etc. $l_i$ is the number of operational sub-dimensions contained in the $i$-th core dimension. (e.g., under the dimension of \textit{Information Content Quality}, the sub-dimensions could be \textit{Information Accuracy and Credibility}, \textit{Scenario Application Completeness}, \textit{User Service Completeness, Scenario Application Relevance} and \textit{User Service Relevance}), etc. The total number of sub-dimensions across all core dimensions is $\sum_{i=1}^L l_i$.
    \item $ M$ is the number of medical scenario categories, defined based on clinical practice and user needs. Examples of scenarios include: \textit{Chronic Disease Consultation in Internal Medicine, Common Disease Consultation in Pediatrics, Postoperative Follow-up in Surgery, Emergency Triage Guidance, and Rare Disease Basic Science Popularization, etc}.
    \item $N$ is the number of medical discipline fields, covering core clinical disciplines. Examples of disciplines include: \textit{General Practice, Internal Medicine, Surgery, Pediatrics, Pharmacy, Nursing, and Emergency Medicine, etc}.
\end{itemize}

\subsubsection{Construction of the 3D Correlation Matrix}
A 3D matrix $\text{Mat}(L \times M \times N)$ is built with "$L$ core dimensions" as rows, "$M$ scenarios" as columns, and "$N$ disciplines" as layers.

\subsection{Training Data Construction}
Based on $\text{Mat}(L \times M \times N)$, structured training data is generated in 5 steps, covering questions, multi-dimensional standards, and multi-quality answers.

\subsubsection{Generation of Question Set $Q$}
\begin{itemize}[left=0pt]
    \item Objective: Generate $Q = \{q_1, q_2, ..., q_{K}\}$ covering $\text{Mat}(L \times M \times N)$, with $K$ determined by scenario complexity.
    \item Process: Traverse $\text{Mat}(L \times M \times N)$; for each $(i, j,k)$, guide an LLM to generate questions via a prompt template. Example: \textit{"Based on \{\{$N_k$\}\} discipline knowledge and for \{\{$M_j$\}\} scenario, design a question that requires evaluating the \{\{$L_{i,x}$\}\} sub-dimension (e.g., medication dosage accuracy) under dimension $L_i$, with sub-dimension compliance verifiable solely from the answer text"}.
\end{itemize}

\subsubsection{Generation of Multi-Dimensional Standards $R_q$ for Question $q$}
For each $q \in Q$, generate the rubric  set $R_q = \{R_{q,1}, R_{q,2}, ..., R_{q,L}\}$ via an LLM (e.g. Qwen). Here $R_{q,i}$ is the standard set from all sub-dimensions for the $i_{th}$ core dimension: $R_{q,i} = \{r_{q,i,1}, r_{q,i,2}, ..., r_{q,i,l_i}\}$, each rubric  is verifiable and quantitatively clear. Example: \textit{For $q$ = "Can a 3-year-old child (in Children scenario) take compound cold medicine while taking acetaminophen (in Medical discipline)?"}, a rubric  for \textit{"drug-drug interaction sub-dimension"} could be \textit{"Must clearly indicate the overlapping acetaminophen risk in compound cold medicine and dose accumulation-induced liver damage."}.
 
\subsubsection{Generation of Multi-Quality Answers}
\textbf{Generation and Scoring for QA Pairs.} 
For each $q \in Q$, generate $Z$ answers of varying quality: $A_q = \{a_q^1, a_q^2, ..., a_q^Z\}$. The sample quality is determined as:
\begin{itemize}[left=0pt]
    \item Poor Answer: Manually perturbed to violate one or more core sub-dimension standards (e.g., omitting drug interaction risks, incorrect doses).
    \item Medium Answers: Generated by a general medical LLM, meeting a middle threshold of $R_q$ sub-dimension standards (e.g., mentioning risks without explaining mechanisms).
    \item Good Answers: Generated by an LLM with authoritative guidelines, meeting a high threshold of $R_q$ sub-dimension standards.
\end{itemize}

For each $qa_q^k = (q, a_q^k)$ ($k=1\sim Z$), multi-dimensional scoring is performed via LLM, outputting an $L$-dimensional score vector:
\begin{itemize}[left=0pt]
    \item Scoring Process: Input $q$, $a_q^k$, and $R_q$ into a strong LLM judge such as GPT4. Score each $R_{q,i}$ sub-dimension on a 1-5 scale (1: Fully non-compliant, 5: Fully compliant), yielding $S_{q,i,k} = \{s_{q,i,1,k}, ..., s_{q,i,l_i,k}\}$.
    \item Core Dimension Score Aggregation: $S_{q,i,k} = \sum_{x=1}^{l_i} (w_{q,i,x} \times s_{q,i,x,k})$, where $w_{q,i,x}$ is the sub-dimension weight under $i_{th}$ core dimension, default to 1.
    \item Final Score Vector: $S_q^k = [S_{q,1,k}, S_{q,2,k}, ..., S_{q,L,k}]$.
\end{itemize}

\subsubsection{Collection of Sample Set}
For each $q$ and $i_{th}$ core dimension, select three candidate samples from $A_q$:
\begin{itemize}[left=0pt]
    \item Good Sample $ab_{i,q}$: answer with $S_{q,i,k} =\mathrm{argmax}_{k \in \{1\sim Z\}} S_{q,i,k}$.
    \item Poor Sample $aw_{i,q}$: answer with $S_{q,i,k} =\mathrm{argmin}_{k \in \{1\sim Z\}} S_{q,i,k}$.
    \item Medium Sample $ar_{i,q}$: random answer with $S_{q,i,k} \in (\mathrm{argmin}_{k \in \{1\sim Z\}} S_{q,i,k}, \mathrm{argmax}_{k \in \{1\sim Z\}} S_{q,i,k})$.
    \item Final Sample Set: $ \mathrm{DS}_q = \{(ab_{1,q}, ar_{1,q}, aw_{1,q}), (ab_{2,q}, ar_{2,q}, aw_{2,q}), ..., (ab_{L,q}, ar_{L,q}, aw_{L,q})\}$.
\end{itemize}

\subsection{SFT Cold-Start Training}
Supervised fine-tuning is performed on the base model using "full-dimensional optimal samples" to achieve initial alignment with medical knowledge and standards.

\textbf{Construction of the SFT Dataset with full-dimensional best sample selection}: For each $q$ and $a_q^k$, calculate total score from all dimensions $\text{S}_{q}^k = \sum_{i=1}^L S_{q,i,k}$, and select the $a_q^k$ with highest $\text{S}_{q}^k$ as best answer $a_q^{best}$ for this question $q$, forming the SFT Dataset $\mathcal{D}_{\text{SFT}}$. The training objective of SFT adopts standard language modeling loss. SFT enables the model to learn medical QA professional expressions, medical logic, and safety requirements, laying the groundwork for RL. Experiments show significant improvements in basic medical knowledge accuracy and safety compared to the base model.

\begin{figure}[t]
    \centering
\includegraphics[width=\linewidth]{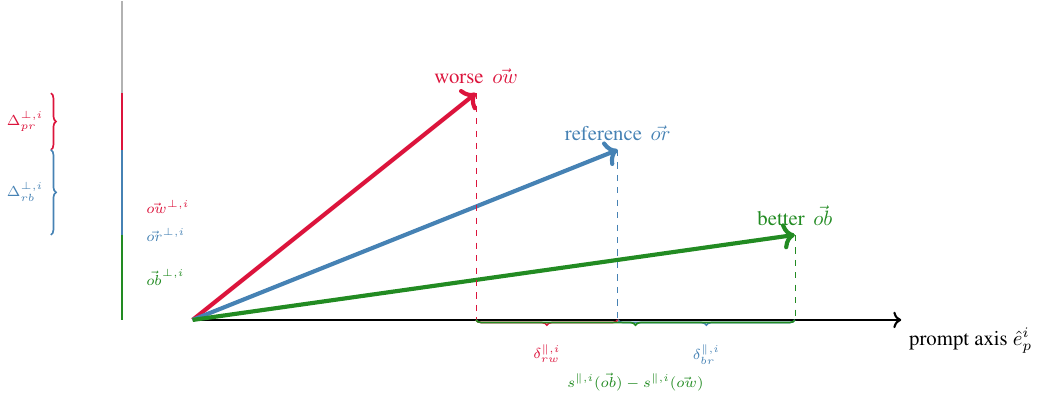}
    \caption{Geometric constraint visualization for reward modeling.}
    \label{fig:reward_toy}
\end{figure}

\subsection{Reward Model Training}
\subsubsection{Construction of Training Samples.} We construct each training sample in $\mathrm{DS}_q$ as a 5-tuple $T_{i,q} = (q, ab_{i,q}, ar_{i,q}, aw_{i,q}, \mathrm{Desc}_{i})$, forming sample set: $T = \{T_{i,q} \mid q \in Q, i\in [1,L]\}$. Here $\mathrm{Desc}_{i}$ is the official description of the $i_{th}$ dimension (e.g., $\mathrm{Desc}_{3}$ = \textit{"Evaluate answer accuracy and safety in medication dosage, indications, and interactions, complying with medical standards"}). A unified multi-dimensional reward model is trained with geometric constraints, inputting $T_{i,q}$ and outputting the $i_{th}$ dimension score (see Fig.~\ref{fig:reward_toy}).

\subsubsection{Model Architecture and Training Process}
\paragraph{(1) Input Representation Encoding}
A shared Transformer-based encoder $\phi(\cdot)$ encodes each element into vector:
\begin{itemize} 
    \item Question–answer vectors: $\vec{ob} = \phi(q\oplus ab_{i,q}) \in \mathbb{R}^d$, $\vec{or} = \phi(q\oplus ar_{i,q}) \in \mathbb{R}^d$, $\vec{ow} = \phi(q\oplus aw_{i,q}) \in \mathbb{R}^d$, where $\oplus$ denotes concatenation. Subscript $\vec{ob}$, $\vec{or}$ and $\vec{ow}$ represent better, reference and worse samples, respectively.
    \item Prompt-anchor vector: $\vec{op}^{\,i} = \phi(q\oplus \text{Desc}_{i}) \in \mathbb{R}^d$, used as the task-dimension axis for projections and ratio consistency in dimension $i$.
\end{itemize}
We use $\vec{op}^{\,i},\vec{ow},\vec{or},\vec{ob}$ as the embeddings for axis-aligned and perpendicular computations relative to the prompt-anchored task-dimension axis.

\paragraph{(2) Prompt-anchored Polar Decomposition}
We construct a polar coordinate system with the prompt as axis. Let $\vec{op}^{\,i},\vec{ow},\vec{or},\vec{ob}$ denote last-token embeddings of prompt (dimension-$i$ axis), worse, reference and better. Define the axis unit $\hat{e}_p^{\,i} = \dfrac{\vec{op}^{\,i}}{\|\vec{op}^{\,i}\| + \varepsilon}$, where $\varepsilon>0$. For any sample vector $\vec{o} \in \mathbb{R}^d$, we decompose it into the axis-aligned scalar projection and the perpendicular component. This polar view makes the training objective geometrically complete: the axis-aligned term $o^{\parallel, i}$ encodes alignment along the task dimension (ensuring $\texttt{worse} < \texttt{reference} < \texttt{better}$ on the prompt axis), while the perpendicular term $\|\vec{o}^{\perp, i}\|$ encodes off-axis deviation (ensuring better/reference are less off-topic than worse). Optimizing both eliminates the unrecognized degree of freedom that appears when only one term is used.
\begin{align*}
    o^{\parallel, i} &= \vec{o}\cdot \hat{e}_p^{\,i},\\
    \vec{o}^{\perp, i} &= \vec{o} - o^{\parallel, i}\, \hat{e}_p^{\,i}
\end{align*}
Here $\hat{e}_p^{\,i}$ is the unit prompt axis for dimension $i$; $o^{\parallel, i}$ is the scalar projection of $\vec{o}$ onto the axis; $\vec{o}^{\perp, i}$ is the component orthogonal to the axis; and axis deviation is measured by $\|\vec{o}^{\perp, i}\|$.
This yields axis similarity via $o^{\parallel, i}$ and axis deviation via $\|\vec{o}^{\perp, i}\|$, jointly determining each sample’s polar placement relative to the prompt.

\paragraph{(3) Axis Similarity Ranking}
We encode the semantic ordering $\texttt{worse} < \texttt{reference} < \texttt{better}$ along the prompt axis using pairwise logistic (Bradley–Terry style) ranking on scalar projections:
\begin{equation*}
    \delta^{\parallel, i}_{br} = \vec{ob}\cdot \hat{e}_p^{\,i} - \vec{or}\cdot \hat{e}_p^{\,i},\quad \delta^{\parallel, i}_{rw} = \vec{or}\cdot \hat{e}_p^{\,i} - \vec{ow}\cdot \hat{e}_p^{\,i},
\end{equation*}
Here $\delta^{\parallel, i}_{br}$ and $\delta^{\parallel, i}_{rw}$ denote axis-margin differences between the pairs (better, reference) and (reference, worse) in dimension $i$.
\begin{equation*}
    L_{\parallel}^{i} = -\log\sigma\big(\delta^{\parallel, i}_{br}\big)\; -\; \log\sigma\big(\delta^{\parallel, i}_{rw}\big).
\end{equation*}
Here $\sigma(\cdot)$ is the logistic function and $L_{\parallel}^{i}$ is the axis-similarity ranking loss for dimension $i$.

\paragraph{(4) Perpendicular Deviation Ranking}
We simultaneously enforce decreasing deviation from the axis: $\texttt{worse} > \texttt{reference} > \texttt{better}$ in $\|\vec{o}^{\perp, i}\|$. Define
\begin{equation*}
    \Delta^{\perp, i}_{wr} = \|\vec{ow}^{\perp, i}\| - \|\vec{or}^{\perp, i}\|,\quad \Delta^{\perp, i}_{rb} = \|\vec{or}^{\perp, i}\| - \|\vec{ob}^{\perp, i}\|,
\end{equation*}
Here $\Delta^{\perp, i}_{wr}$ and $\Delta^{\perp, i}_{rb}$ denote the margins in perpendicular deviation between (worse, reference) and (reference, better) in dimension $i$.
\begin{equation*}
    L_{\perp}^{i} = -\log\sigma\big(\Delta^{\perp, i}_{wr}\big)\; -\; \log\sigma\big(\Delta^{\perp, i}_{rb}\big).
\end{equation*}
Here $L_{\perp}^{i}$ is the perpendicular-deviation ranking loss for dimension $i$.
This ranking encourages reference/better to align closer to the prompt axis than worse.

\paragraph{(5) Total Objective}
The final objective balances axis similarity and deviation control:
\begin{equation*}
    \mathcal{L}_{\text{RM}}^{i} = \alpha\, L_{\parallel}^{i} + (1-\alpha)\, L_{\perp}^{i},\quad \alpha \in (0,1).
\end{equation*}
Here $\mathcal{L}_{\text{RM}}$ is the total objective and $\alpha$ balances axis similarity and deviation control. The prompt-anchored polar view separates “along-axis semantics” from “off-axis deviation”, yielding interpretable gradients and robust geometry. Stability is ensured by saturating $-\log\sigma(\cdot)$ on large margins and unit-axis normalization and $\varepsilon$ safeguards in norms.

\begin{figure}[t]
    \centering
\includegraphics[width=\linewidth]{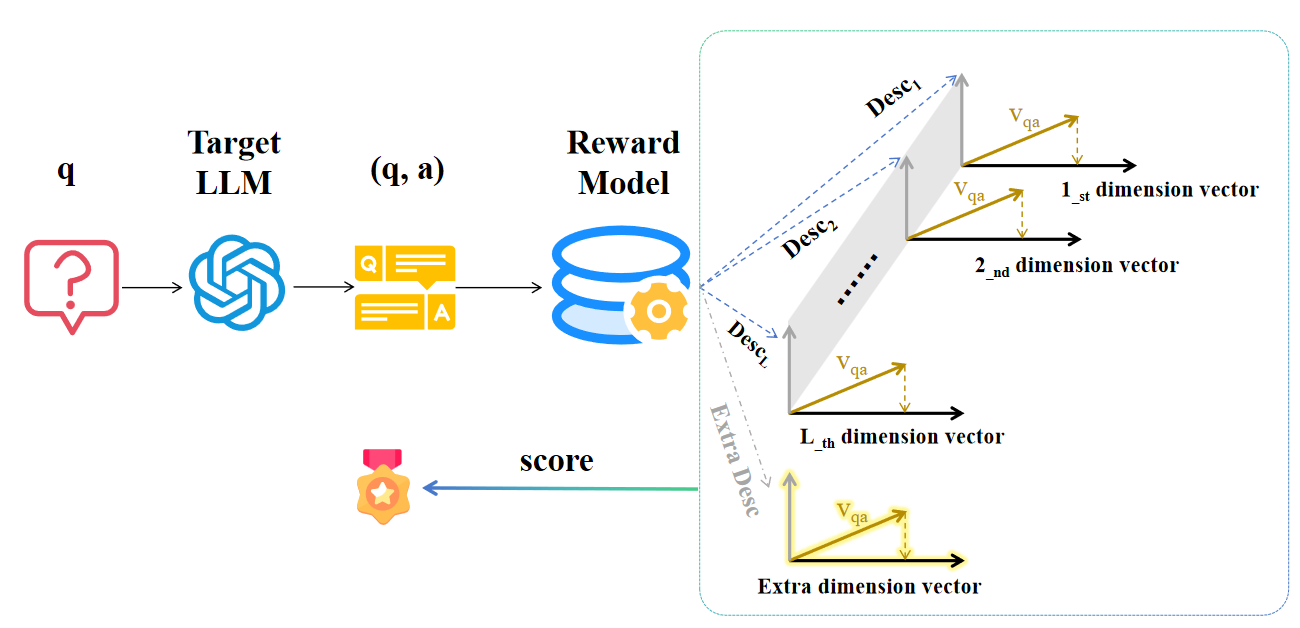}
    \caption{Scoring with a unified reward model from multidimensions during reinforcement learning.}
    \label{fig:scoring}
\end{figure}

\subsection{Reinforcement Learning}
Our framework can be adapted to various reinforcement learning methods, such as GRPO (adopted in this paper) \citep{shao2024deepseekmath} or DAPO \citep{yu2025dapo}, with scores from reward model trained via MR-RML as reward signals to optimize the SFT-enhanced model.

\textbf{Online Data Generation.} For each $q \in Q$, the SFT-enhanced model ($\text{Policy}_0$) generates $G$ online answers, forming $\text{QA}_{\text{online},q} = \{(q,a_q^{1,\text{on}}), ..., (q,a_q^{G,\text{on}})\}$.

\textbf{Reward Model Scoring and Overall Score Calculation.} As shown in Figure~\ref{fig:scoring}, for each $(q,a_q^{g,\text{on}})$ in $\text{QA}_{\text{online},q}$, given input $[q, a_q^{g,\text{on}}, \text{Desc}_{i}]$ into RM for each $L_i$, yielding dimension score $s_{i}^g$, and aggregate them into a overall score $S_{\text{on}}^g = \sum_{i=1}^{L}\cdot s_{i}^g$.


\begin{figure}[t]
    \centering
\includegraphics[width=\linewidth]{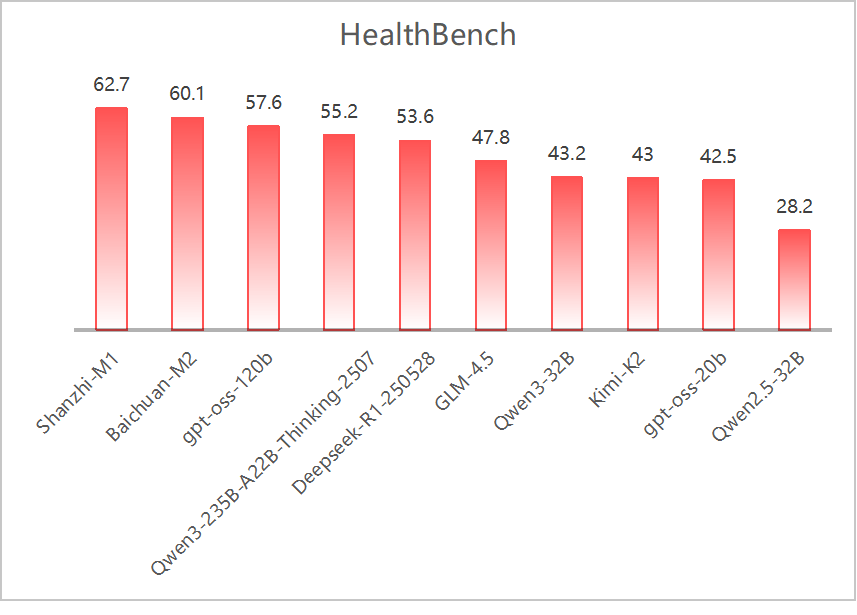}
    \caption{Overall performance on HealthBench (Full) for open-sourse LLMs.}
    \label{fig:hb_all}
\end{figure}

\begin{figure}[t]
    \centering
\includegraphics[width=\linewidth]{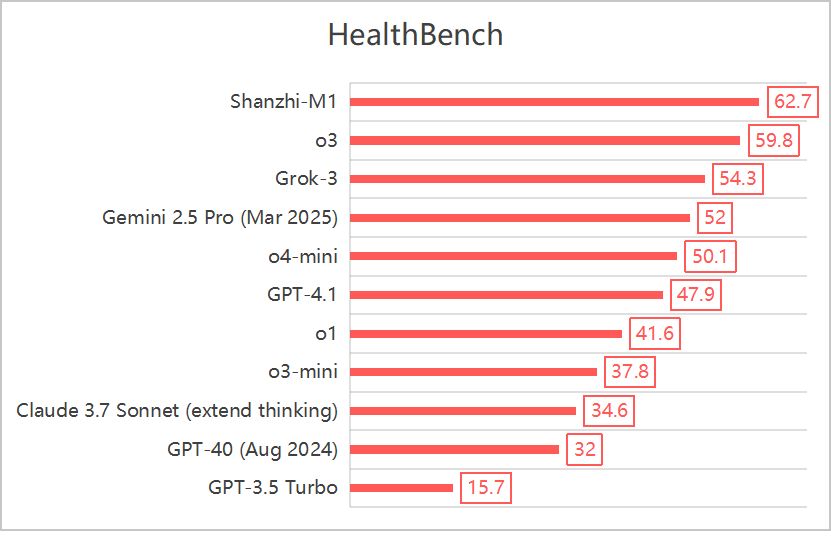}
    \caption{Overall performance on HealthBench (Full) for closed-sourse LLMs.}
    \label{fig:closed_all}
\end{figure}

\begin{figure}[t]
    \centering
\includegraphics[width=\linewidth]{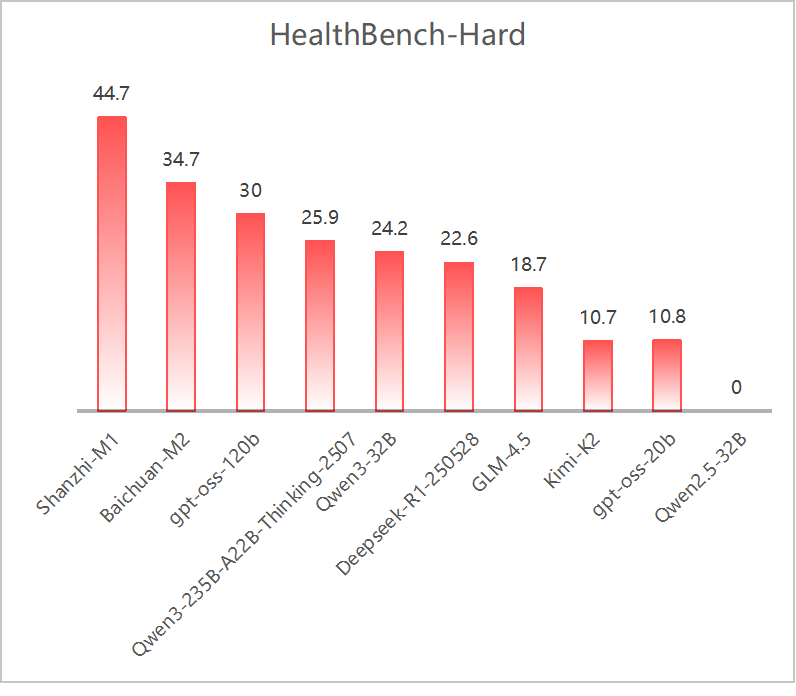}
    \caption{Performance on HealthBench Hard (high-complexity tasks) for open-sourse LLMs.}
    \label{fig:hb_hard}
\end{figure}

\begin{figure}[t]
    \centering
\includegraphics[width=\linewidth]{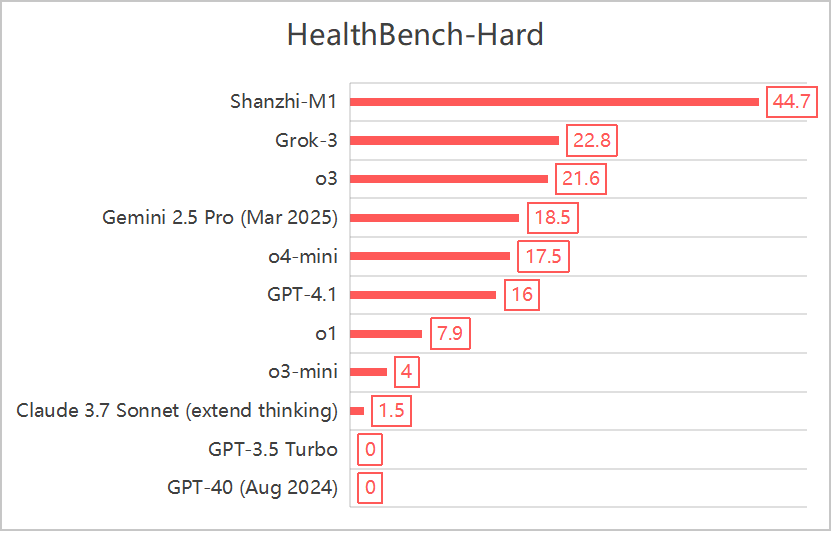}
    \caption{Performance on HealthBench Hard (high-complexity tasks) for closed-sourse LLMs.}
    \label{fig:closed_hard}
\end{figure}

 \section{Experiments}
\label{sec:experiments}

To rigorously validate the effectiveness of the proposed MR-RML paradigm in enhancing the clinical usability and generalization of medical large language models (LLMs), we conduct comprehensive evaluations on HealthBench—a gold-standard healthcare benchmark. Our experiments are designed around three core objectives: (1) measuring overall performance against state-of-the-art models, (2) assessing robustness on high-complexity tasks, and (3) verifying the paradigm's advantages in clinical scenario alignment and deployment efficiency. All evaluation processes strictly follow those in \citep{arora2025healthbench}, with the overall score reported in percentage (the original score has been converted to a 100-point scale).

\subsection{Evaluation Benchmark: HealthBench}
\label{subsec:benchmark}

We adopt HealthBench \citep{arora2025healthbench}, a large-scale, clinically grounded evaluation suite, as the primary testbed. This benchmark is designed to simulate real-world medical interactions and bridges the gap between technical evaluation and clinical practice, making it highly compatible with the goals of our MR-RML paradigm. All training data is distinct from the data included in the evaluation benchmark HealthBench. Key features of HealthBench include:
\begin{itemize}[left=0pt]
    \item \textbf{Data Scale}: 5,000 multi-turn conversational question-answering pairs .
    \item \textbf{Evaluation Rigor}: 48,562 granular rubric criteria developed by 262 board-certified physicians, aligned with authoritative clinical guidelines.
    \item \textbf{We conducted on 2 specialized Subsets}:
    \begin{itemize}
        \item HealthBench (Full): Assesses overall clinical performance.
        \item HealthBench Hard: 1,000 high-complexity questions involving cross-lingual inputs and dual perspectives.
    \end{itemize}
\end{itemize}

\subsection{Experimental Setup}
\label{subsec:setup}

\subsubsection{Baseline Models}
\label{subsubsec:baselines}

Our show out model $\text{MR-RML}_{qwen32b}$ (short for $\text{Shanzhi-M1}$) performance trained on Qwen3-32B using the MR-RML paradigm. We compare it against various leading open-source and closed-source LLMs to ensure a comprehensive evaluation:
\begin{itemize}[left=0pt]
    \item \textbf{Open-Source Models}: Baichuan M2 \citep{dou2025baichuan}, gpt-oss-20b and 120B \citep{agarwal2025gpt}, Qwen3-235B-A22B \citep{yang2025qwen3}, DeepSeekR1 \citep{guo2025deepseek}, GLM-4.5 \citep{zeng2025glm}, Qwen3-32B\citep{yang2025qwen3}, Kimi-K2 \citep{team2025kimi}, Qwen2.5-32B \citep{team2024qwen2}.
    \item \textbf{Closed-Source Models}: OpenAI O3 \citep{yan2025comparative}, Grok 3 \citep{mohammed2025deepseek}, Gemini 2.5 Pro \citep{comanici2025gemini}, Claude 3.7 \citep{joshi2025open} and GPT series \citep{nakajima2024comparison, achiam2023gpt}.
\end{itemize}

Consistent with HealthBench's official protocol, the primary metric is the \textit{overall rubric compliance score} (0–100), which aggregates scores across all clinical criteria. 

\subsection{Experimental Results}
\label{subsec:results}

\textbf{Overall Performance on HealthBench.}
As illustrated in Figure~\ref{fig:hb_all} and Figure~\ref{fig:closed_all}, our model achieves a leading overall score of 62.7 on the full HealthBench evaluation, outperforming all open-source models and most closed-source counterparts. Specifically:
\begin{itemize}[left=0pt]
    \item It surpasses the top open-source competitor (62.7) by a marginal gap while maintaining a significant lead over other open-source models.
    \item Among closed-source models, Our model outperforms O3 (59.8), Grok 3, Gemini 2.5 Pro, and GPT-4.1, demonstrating strong competitiveness.
\end{itemize}

\textbf{Performance on HealthBench Hard.}
HealthBench Hard tests models' ability to handle complex, real-world clinical challenges. As shown in Figure~\ref{fig:hb_hard} and Figure~\ref{fig:closed_hard}, Our model $\text{Shanzhi-M1}$ achieves a score of 44.7, making it one of only two models globally (alongside GPT-5) to exceed the 40-point threshold.

\subsubsection{Scenario-Specific performance}
\label{subsubsec:scenario}

$\text{Shanzhi-M1}$ leads in five core clinical scenarios (see Table \ref{tab:clinical_scenarios}), directly validating the effectiveness of the MR-RML paradigm.
\begin{table}[htbp]
\centering
\caption{Performance in Five Core Clinical Scenarios}
\label{tab:clinical_scenarios}
\begin{tabular}{|l|c|}
\hline
\textbf{Clinical Scenario} & \textbf{Score} \\
\hline
Emergency Referrals & 74.3 \\
Communication & 69.6 \\
Context Awareness & 52.4 \\
Context Seeking & 58.5 \\
Global Health & 59.2 \\
\hline
\end{tabular}
\end{table}

The implementation of the MR-RML framework achieved a groundbreaking over ninety percent reduction in expert labor costs. This was realized by streamlining the annotation process: experts only need to participate in defining and validating core clinical standards (aided by LLMs), eliminating the need for large-scale manual sample labeling. As a result, human resource input is reduced to less than one-tenth of the original, while maintaining stable and comparable clinical application effectiveness—this labor-cost-efficient approach does not compromise performance. Additionally, the performance enabled by geometric projection reference constraints is significantly superior to that of the conventional reward model learning method.

\section{Conclusion}
This study addresses the fundamental limitations of current medical LLM alignment paradigms, which fail to reconcile medical cognitive characteristics, dynamic standard adaptation, and cost-effective multi-dimensional optimization. We propose MR-RML, a comprehensive framework that reconstructs the alignment pathway of medical LLMs through medical standard integration, multi-dimensional reward modeling, and geometric constraint-based optimization.
MR-RML’s core contributions are threefold. First, the 3D medical standard system (Dimensions-Scenarios-Disciplines) resolves the long-standing disconnection between evaluation metrics and clinical practice by embedding authoritative medical standards into data generation and full training flow. Second, the independent multi-dimensional reward model enables refined decomposition of clinical evaluation criteria, replacing high-cost real-time rubric scoring with internalized reward signals—striking a balance between professional consistency and deployment efficiency. Third, the geometric projection reference constraint transforms medical cognitive logic into mathematical regularization, ensuring scoring gradients align with clinical reasoning and reducing reliance on scarce high-quality expert annotations via synthetic data.
Experimental results on HealthBench confirm MR-RML’s effectiveness: the model outperforms most state-of-the-art open and closed-source LLMs on overall clinical performance, achieves leading results on high-complexity tasks (HealthBench Hard), and excels in critical scenarios such as emergency referrals and medical communication. These findings validate that MR-RML successfully addresses the three core bottlenecks of medical LLM alignment, providing a scalable, standard-extensible, and cost-controllable solution for clinical AI deployment.
Future work will focus on three directions: (1) expanding the 3D medical standard system to cover more specialized disciplines (e.g., radiology, pathology) and regional clinical guidelines; (2) integrating multi-modal medical data (e.g., imaging, lab results) into the reward modeling framework to support comprehensive clinical decision-making; (3) optimizing the geometric constraint loss function for real-time adaptation to dynamic guideline updates, further enhancing the framework’s practical clinical utility. Ultimately, MR-RML aims to accelerate the translation of medical AI from technical demonstrations to trustworthy clinical tools that improve care quality and accessibility.

\section{Future Work}
While the MR-RML framework demonstrates strong performance, future work should focus on broader benchmarking across diverse medical benchmarks to enhance generalizability, multimodal integration of imaging or lab data into reward modeling for comprehensive decision support, and dynamic adaptation of geometric constraints for real-time guideline updates to improve clinical utility, ensuring continuous alignment with evolving medical standards.
 
\bibliographystyle{acl_natbib}
\bibliography{anthology}

\appendix


\end{document}